\documentclass[10pt,twocolumn,letterpaper]{article}

\usepackage[pagenumbers]{cvpr} 

\usepackage{graphicx}
\usepackage{amsmath}
\usepackage{amssymb}
\usepackage{booktabs}
\usepackage[pagebackref,breaklinks,colorlinks]{hyperref}

\usepackage{amssymb}
\usepackage{latexsym}

\usepackage{url}
\usepackage{xcolor}

\usepackage{hyperref}

\usepackage{subcaption}
\usepackage{makecell}

\usepackage{tabularx}
\usepackage{graphicx}
\usepackage{arydshln} 

\usepackage{diagbox} 
\usepackage{multirow}

\definecolor{newcolor}{rgb}{.8,.349,.1}

\usepackage[capitalize]{cleveref}
\crefname{section}{Sec.}{Secs.}
\Crefname{section}{Section}{Sections}
\Crefname{table}{Table}{Tables}
\crefname{table}{Tab.}{Tabs.}

\def\cvprPaperID{*****}
\def\confName{CVPR}
\def\confYear{2022}

\begin{document}

\title{Towards General Purpose Medical AI: Continual Learning \\ Medical Foundation Model}

\author{Huahui Yi$^{1~*}$, Ziyuan Qin$^{1~*}$, Qicheng Lao$^{2,5}$ \thanks{Equal contribution ~~~~~ $^{\dagger}$Corresponding author}~~$^{\dagger}$, Wei Xu$^{1}$, Zekun Jiang$^{1}$, Dequan Wang$^{3,5}$, \\ Shaoting Zhang$^{5}$, Kang Li$^{1,4,5~{\dagger}}$\\
$^1$West China Biomedical Big Data Center, West China Hospital, Sichuan University\\
$^2$School of Artificial Intelligence, Beijing University of Posts and Telecommunications, China\\
$^3$Shanghai Jiao Tong University, Shanghai, China\\
$^4$Sichuan University Pittsburgh Institute, Chengdu, China\\
$^5$Shanghai Artificial Intelligence Laboratory, Shanghai, China \\
}

\maketitle

\begin{abstract}

Inevitable domain and task discrepancies in real-world scenarios can impair the generalization performance of the pre-trained deep models for medical data. Therefore, we audaciously propose that we should build a general-purpose medical AI system that can be seamlessly adapted to downstream domains/tasks. Since the domain/task adaption procedures usually involve additional labeling work for the target data, designing a data-efficient adaption algorithm is desired to save the cost of transferring the learned knowledge. Our recent work~\cite{qin2022medical} found that vision-language models (VLMs) are efficient learners with extraordinary cross-domain ability. Therefore, in this work, we further explore the possibility of leveraging pre-trained VLMs as medical foundation models for building general-purpose medical AI, where we thoroughly investigate three machine-learning paradigms, \ie, domain/task-specialized learning, joint learning, and continual learning, for training the VLMs and evaluate their generalization performance on cross-domain and cross-task test sets. To alleviate the catastrophic forgetting during sequential training, we employ rehearsal learning and receive a sharp boost in terms of generalization capability. In a nutshell, our empirical evidence suggests that continual learning may be a practical and efficient learning paradigm for the medical foundation model. And we hope researchers can use our empirical evidence as basement to further explore the path toward medical foundation model.\footnote{work in progress.}

\end{abstract}

\begin{figure}[htbp]
\centering
\includegraphics[width=0.47\textwidth]{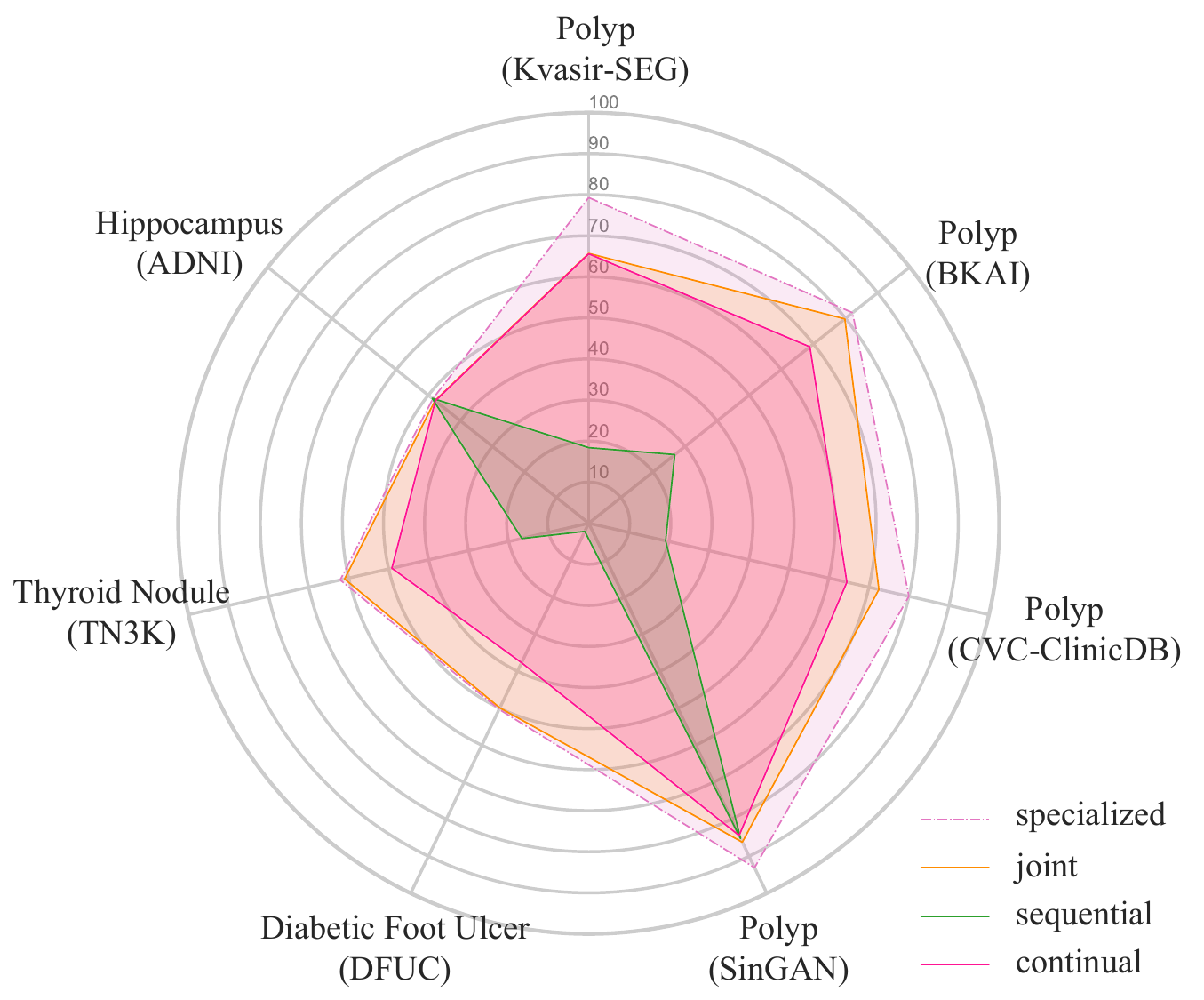}
\caption{Results on three different paradigms for implementing universal medical foundation models: specialized learning, joint learning, and continual learning. Continual learning may be a practical and efficient paradigm for the medical foundation model.} 
\label{fig:fig1}
\end{figure}

\section{Introduction} 
\label{sec_intro}
Recently, foundation models have gained a lot of popularity in deep learning. People use these huge pre-trained models with billions or even trillion parameters to solve downstream specific tasks such as machine translation and document classification~\cite{weng2020acquiring,lee2019patentbert}. Large language models (LLM) are the prototype of foundation models, which are trained on a large corpus collected from the internet with unsupervised tasks such as masked language modeling (MLM)~\cite{devlin2018bert}. Foundation models have been prevalent in natural language processing fields for a while in recent years, but it took years for this trend to deep vision learning and its related fields. Compared to LLMs, foundation models for vision tasks have two possible pre-training approaches. The first approach is to reconstruct the input image using masked auto-encoders (MAE)~\cite{he2022masked} directly, and the second is to introduce the pairing text descriptions of images as weak supervising labels. The first approach has shown its extraordinary imaging understanding capability in many downstream vision tasks~\cite{bachmann2022multimae,dong2022bootstrapped}. However, based on some recent research, the MAE method may not have strong domain generalization capability compared to the second approach. The models trained with the second approach are called vision-language models (VLM), and they have promising generalization capability for unseen concepts. Inspired by the development of foundation models and the success of VLMs, we naturally think of applying natural image foundation models to the future of medical imaging models, as the medical imaging domain urgently needs large models with generalization capability.

Although deep learning algorithms have been progressing greatly in medical image analysis tasks, from classification to target area detection/segmentation, a deep model pre-trained on a specific dataset usually suffers from the domain/task shift problem when facing a new incoming dataset with different distribution. Due to this problem, applying a pre-trained medical imaging model to target data collected from other sources is not a seamless procedure. On the other hand, medical machine learning practitioners tend to develop task-specialized models that can only be applied to a very specific medical task (\eg, classifying a rare type of tumor), and those models, undoubtedly, are also trained with a limited size of data. Thus, one may need to train an ad-hoc model for different target data or tasks, and this process usually involves additional labeling work and requires the data in advance for fine-tuning, which is not applicable for real-time use cases.  This relatively cumbersome process has been a hurdle for adopting a deep-learning-based image analysis system in practice. 

In realistic scenarios, the incoming data may come with all sorts of variations from acquisition to processing. 
Unfortunately, we have no control over the aforementioned variables in real-world environments, and acquiring the target data in advance is also unrealistic in many scenarios, especially for real-time clinical usage. Therefore, these issues suggest we need to propose new paradigms to enhance the domain/task generalization capability of our deep models regarding the real-time incoming data. \textbf{Therefore, we envision there is a new paradigm for building a general-purpose medical artificial intelligence (AI) model which can easily be adapted for various medical tasks and domains. And, as we mentioned before, building medical image specialized foundation models may be the correct path toward general-purpose medical AI. } 

As one of the pioneering work pushing toward this vision, we can see there is still a huge vacancy left for research and development to build a general-purpose AI system, but a foundation model may be the answer to our goal. And in our recent series of studies, we have the very first taste of how to build a foundation model for medical AI systems efficiently.
Our previous work~\cite{qin2022medical} showed that large-scale pre-trained vision language models have such generalization capabilities in the new domains (\eg, from the natural image domain to the medical domain) by activating the knowledge with appropriate text prompts. In that work, we proposed multiple prompt generation approaches, which can either artificially or automatically generate text prompts. These prompts contain the expressive attributes described in the target concept to trigger the aligned visual representation acquired during the pre-training stage in a new domain. And the effectiveness of our method has been supported by excessive experiments and the superior performance we achieved in many public medical image datasets. Therefore, from the lessons we learned in the above work, we assume the strong cross-domain stability of VLM not only exists in prompt-based learning but also exists in training with heterogeneous data and general-purpose tasks, approaching the vision of medical foundation models.

In this work, we thoroughly examine the generalization capability of the VLMs trained with our newly proposed paradigms and explore the possibility of leveraging pre-trained VLMs as medical foundation models for building general-purpose medical AI. The first learning paradigm we evaluate is domain/task-specialized learning, \ie, to train a model on a specific domain/task and test its generalization performance on unseen domains/tasks. With the unsatisfying results, we then move on to the joint learning paradigm which feeds the models with heterogeneous data for general purposes. Although the cross-domain or cross-task generalization results are acceptable compared to specialized learning, jointly training such a model usually requires accessing the target data in advance, which is not applicable in the real world. Therefore, we finally evaluate the continual learning paradigm to train the VLMs on sequentially arriving data. By only including a small size of replay buffer for rehearsal, the catastrophic forgetting can be efficiently alleviated and thus improving the models' generalization. The overall performance pattern for the above learning paradigms is visualized in Fig.~\ref{fig:fig1}. We can see that both joint learning and continual learning show promising generalization performance among different domains of polyp detection data and among different medical tasks (hippocampus, thyroid nodule, diabetic foot ulcer, and polyp detection).   

The main contribution of this work can be concluded as follows:
1) We thoroughly evaluate the vision-language models under three learning paradigms for their cross-domain and cross-task generalization capability;
2) We propose to use the continual learning paradigm as an efficient and practical way to build medical foundation models toward general-purpose medical AI;
3) We find the catastrophic forgetting problem is a vital problem that undermines the models' generalization capability, which can be efficiently restored by employing rehearsal learning.

\section{Related Work}
\label{sec_related}

\subsection{Foundation models in healthcare} 
Foundation models have gained popularity and become a new paradigm for AI. When we say foundation model, we usually refer to those models which are pre-trained with large-scale heterogeneous data and can be transferred to various downstream tasks. The related review studies have discussed the opportunities and risks of the foundation model~\cite{bommasani2021opportunities} and foundation model in healthcare~\cite{wojcik2022foundation}, where the foundation model was labeled as the ``new AI". The intelligence exhibited by the foundation model is mainly manifested by emergence and homogenization capability, which is not available in past AI models, providing a new direction for general AI~\cite{bommasani2021opportunities}. Emergence is mainly demonstrated by the ability to in-context learning and to adapt to different downstream tasks, while homogenization lies in the ability to process different heterogeneous data and can be transferred to a broader range of tasks. By any measure, the foundation model brings more convenience to AI practices, including in healthcare. Medical AI has always been limited by the size of clinical data and has struggled to go deeper. The medical foundation model may be a game-breaker for medical AI by learning from a wide range of medical data and being transferred to data-limited downstream tasks. Currently, it is worth exploring how to design the datasets and learning strategies needed to build the medical foundation model. In our study, we do some explorations in this direction and give some problem summaries.

\subsection{Prompting for foundation models} 
Recently, prompt based model inference and knowledge extraction has been widely adopted in NLP field~\cite{lama}. People find that prompts can activate the knowledge contained in LLMs without tuning their parameters~\cite{Schick2020ExploitingCF}. Thus, we follow this line of works~\cite{lama, HaoyuSong2022CLIPMA, JianweiYang2022UnifiedCL} to design the our model generation method. Along with the development of pre-trained VLMs~\cite{ALIGN, CLIP}, prompt learning become more and more important in tasks such as text-image retrieval and image classification through template prompts. In our last work~\cite{qin2022medical}, we mainly focus on leveraging the prompt text to activate pre-trained VLMs knowledge of unseen objects, and show excellent performance on various public medical datasets across different domains and tasks. 

\subsection{Joint learning and continual learning} 
The foundation model has emerged to better adapt to different downstream tasks that can span different imaging modalities, disease domains, and even processing tasks. There have been many strategies for how to make neural networks learn to multitask knowledge. Joint training/learning is a training strategy directing joint training of datasets with different domains and tasks, which has been reported for successfully working in image processing and language modeling~\cite{tompson2014joint,Zhang2019JointLO}. 

In addition, continual learning (CL) provides another training strategy for learning  from different task data, which also be called  incremental learning or life-long learning. Neural networks often suffer from catastrophic forgetting~\cite{french1999catastrophic} when learning different data again and again, i.e., forgetting what they learned from the previous data, and this limits deep learning models to handle different downstream tasks. Continual learning tries to solve this problem and usually includes the following implementations: context-specialized components, model regularization, feature replay~\cite{Lao2021ATC}, and dynamic network expansion, et al~\cite{thengane2022clip,van2022three}. The continual learning scenarios include class-incremental, domain-incremental, and task-incremental learning for both classification and generation tasks~\cite{van2022three, Lao2020FoCLFC}, and it is precisely this need for the foundation model to learn as much knowledge as possible and migrate them to across-class, across-domain, and across-task datasets for efficient reasoning. Recent studies have demonstrated that transformer can do continual learning~\cite{douillard2022dytox} and Contrastive Language-Image Pretraining (CLIP) model as a VLM is an efficient continual learner~\cite{thengane2022clip}, which all provide ideas for our research.

\begin{figure*}[htbp]
\centering
\includegraphics[width=1\textwidth]{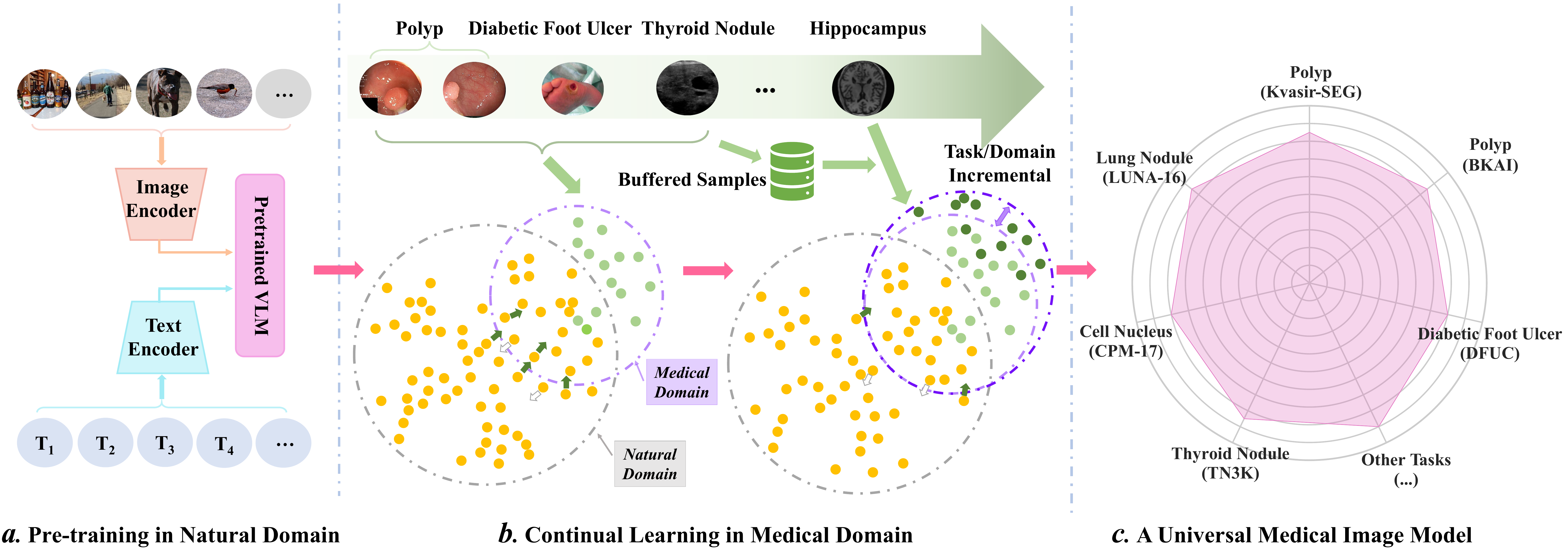}
\caption{Overview of the roadmap to universal medical foundation model. The roadmap is made up of three steps: \textbf{a.} model is pre-trained on a huge number of image-text pairs in the natural domain to obtain generic concepts which can be transferred to the medical domain; \textbf{b.} incremental extension of the medical domain/task concepts through continual learning applicable to real-life medical scenarios; \textbf{c.} the ideal model will have a balanced performance across a range of medical domains and tasks and generalizable to unseen domains/tasks.} 
\label{fig:fig2}
\end{figure*}

\section{Methods}
\label{sec_method}
In this section, we first briefly discuss how to apply VLMs to different medical image analysis tasks in zero/few-shot manners with text prompts. And we will discuss how to leverage LLMs as knowledge source to generate our text prompts. Then we will move toward the first training paradigm we proposed, in which we train the VLMs with heterogeneous data and different medical image understanding tasks jointly. Ultimately, We will explain our two continual learning implementations and discuss the catastrophic forgetting problem and our solution.

\subsection{Pre-trained VLMs}
Vision-language models can learn from the massive online image-text paired data using aligned language information as supervision labels. These models usually have separate text and visual encoders that are aligned semantically in a joint embedding space, as shown in Fig.~\ref{fig:fig2}(a). Since the language descriptions of a concept are typically consistent across various data domains, using language prompts to substitute the class labels gives the VLMs impressive zero-shot vision task performance. Grounded Language-Image Pre-Training (GLIP)~\cite{li2022grounded} is a recently proposed model which reformulates object detection as phrase grounding tasks. GLIP takes the text phrase prompts instead of class labels to supervise the grounding module. And the model will learn to align the grounded visual area with the regarding text phrase in the prompts during the training process, and, therefore, GLIP can semantically align the visual representation and the concept in a joint embedding space. To be specific, the above process can be formulated as the following:
\begin{multline}
O=Enc_I(\text{Image}), P=Enc_L(\text{Prompt}),\\
S_{ground} = OP^\top, L_{cls}=Loss(S_{ground};T),
\end{multline}
where $ O \in \mathbb{R}^{N \times d}, P \in \mathbb{R}^{M \times d} $ denote the image and text features respectively, $ S_{ground} \in \mathbb{R}^{N \times M}$ represents the cross-modal alignment scores, and $T \in \{0,1\}^{N \times M}$ is the target matrix.

\subsection{Adapting VLMs to the medical domain through text prompts}
In our last work~\cite{qin2022medical}, we proposed multiple methods for prompt generation, including manual and automatic approaches. The essential idea of prompts designation is to inject prior knowledge of the target concept/object into the prompts. To be more specific, we need knowledge about the visual appearance of a target object. We show that such knowledge can more efficiently activate the pre-trained VLMs and the related visual representation with which the text prompts are aligned.
Therefore, we propose several methods to gain such knowledge from different knowledge sources. The manual method is straightforward, but it requires much effort for human annotators to find useful prompts. So we decide to automize this process by leveraging the knowledge contained in LLMs or VLMs. To elicit knowledge from the LLMs, we make the specialized medical LLMs do the masked token prediction task by giving a template sentence. We use the masked token to represent the attribute desired adjective in the template sentence. For example, if we want to know the color attribute of polyps, we then feed a template -- `The color of polyps is [masked]' to the LLMs and let the LLMs predict the masked token. The predicted word should be the attribute value we are looking for. As for extracting knowledge from the VLMs, we simply find a pre-trained VQA model~\cite{OFA} to answer our question regarding a given image. And the question is, of course, about the attribute we need.  

\subsection{Exploration of medical foundation model}
After we show the amazing generalization capability of the pre-trained VLMs from the natural image domains to medical domains, we further investigate whether such capability exists in the process of adapting different medical domains and tasks. In this work, we design several approaches to verify the generalization power of the VLMs in a range of data domains and medical tasks. Therefore, we trained a series of domain/task-specialized models. As our preliminary experiment results suggest, neither domain nor task-specialized pre-trained models show acceptable cross-domain or cross-task performance. Therefore, we adopt two advanced methods to make the very first step on the path toward medical foundation models. The first method allows us jointly train our large models with mixed data from different domains and tasks. Furthermore, the second method is more practical in the real world. We explain these two methods in detail in the following passages.

\subsubsection{Generalization through joint learning}
Joint learning usually refers to an approach aiming to learn from a range of datasets at once instead of training an independent model on one dataset each time. Suppose we are given n datasets from different domains/tasks, and we want to train a neural network M to learn the patterns contained in the data. Instead of training n models on each dataset separately, we want to use just one model to simultaneously capture the patterns across various domains or tasks. To achieve this goal, we optimize the parameters through a unified training paradigm for different datasets to let our models learn the patterns across domains/tasks jointly. 

Formally, we are given $n$ different datasets $D=\{(x_i, y_i)\}^n_{i=1}$, and each one of them is from a separate domain/task. Then we train a model $M$ to approach the joint distribution of $P ~ (x_i, y_i) $. This process can be formalized as follows:
\begin{equation}
    P_{\theta}(x, y) = \frac{1}{n}\sum_{n=1}^{n}\theta(x=x_i, y=y_i),
\end{equation}

The joint dataset $D$ is more representative because the data come from different domains/tasks can better reflect the unknown original data distribution. 
 
And that's the essential motivation driving us to train a model for all domains/tasks instead of training several domain/task-specialized models. As mentioned above, we will use experiments to show that domain/task-specialized models' generalization performance is not comparable with the joint-training large models. 
Though we've discussed so many advantages of joint learning, some remaining problems still deter the joint learning method from becoming the only correct path toward the medical foundation models. Firstly, in clinical scenarios, target-domain/task data is not accessible in advance. So they are not available for joint training, and we have to finetune them accordingly. 

However, in this process, we may encounter the catastrophic forgetting problem, which means the pre-trained model forgets the acquired knowledge during the pre-training stage and, thus, lose the generalization capability. 
Secondly, Training a large-size pre-trained model may not be practical in real-world circumstances, considering the sensitivity and privacy of the data. Thus, we may only start training our model with a relatively small-size and homogeneous data distribution at the beginning. All of the above problems point to another machine learning paradigm and a natural option -- continual learning.

\subsubsection{Generalization through continual learning}
Continual learning, also known as lifelong learning, is a machine learning paradigm designed for learning patterns from a continuous data stream, rather than learning from a fixed dataset all at once. The common interest of continual learning and domain generalization is to find a way that can keep the acquired knowledge and keep learning from incoming data. As we pointed out before, one of the major challenges of applying a pre-trained model for medical image analysis tasks is that we have no control over the quality of incoming data. Thus, even if we could have enough labeled data to finetune our pre-trained model, we still have to prevent forgetting the knowledge acquired previously during the finetuning process. So we think CL could be an ideal solution for such worries. In fact, there are two subfields of CL called task-incremental CL and domain-incremental CL dedicated on solving the problem we just mentioned. 

In this work, we borrow the idea of task-incremental CL and domain-incremental CL to train our models and verify the generalization capability of our model. As Fig.~\ref{fig:fig1}(b) illustrated, we first train a model with data $D_i$ from only one domain/task, and we keep increasing data from different domains/tasks during the continual learning process to test the model's performance. We apply two different methods of continual learning to train our model and observe the forgetting patterns. The first method is called sequential learning, which is the fundamental benchmark of continual learning. To be specific, we start by training the model with data $D_i$ from a sole domain. Then we continually train the model by feeding data $D_{i+1}, D_{i+2}, ..., D_n$ from incoming domains. We test the model's performance after each training with one domain on all of the test sets to see whether the model is enduring catastrophic forgetting problems. The other method is called rehearsal or replay learning, a popular and classical training method adopted in continual learning. The training process of rehearsal learning is pretty much the same as sequential learning, except for a small but vital change. This change is keeping a replay buffer $B = \{x_i^j, y_i^j |  i\in \Vert D \Vert, j\in n\}$ that holds some data and labels randomly sampled from the previous datasets, and the data and labels in the replay buffer will be periodically fed to the model even if it's learning the target domain's data. Though there are some more sophisticated prioritized sampling strategies rather than random sampling, we still observe a huge improvement in terms of alleviating the catastrophic forgetting problem by introducing this simple strategy. We will analyze this phenomenon in detail in the experiments section. For the task-incremental CL, we use the same training methods, except we add data from another task at each time.

\section{Experiments}

\begin{table*}[ht]
\centering

\caption{Cross-domain generalization performance on the polyp datasets (AP\%)}
\label{tab1}
\renewcommand\arraystretch{1.5}
\setlength{\tabcolsep}{0.9mm}{
\begin{tabular}{l|ccccccccc}
\toprule

\multirow{2}{*}{\diagbox[width=7em,height=3.63em]{Train}{Test}} & \multicolumn{8}{c}{Polyp} & \multirow{2}{*}{\fcolorbox{lightgray}{lightgray}{AVG}}  \\ 
\cmidrule(lr){2-9}
 & BKAI   & CVC-ClinicDB   & Kvasir-SEG    & Kvasir-Sessile & SinGAN        & CVC-300       & CVC-ColonDB   & ETIS &  \\ \hline

BKAI           & \textbf{79.3} & 65.8           & 69.1          & 60.8           & 64.4          & 71.5          & 48.5          & 61.8  & \fcolorbox{lightgray}{lightgray}{65.2} \\
CVC-ClinicDB   & 64.5          & \textbf{82.2}  & 61.6          & 60.4           & 62.8          & \textbf{72.4} & \textbf{61.1} & 53.2  & \fcolorbox{lightgray}{lightgray}{64.8} \\
Kvasir-SEG     & 62.1          & 59.6           & \textbf{75.9} & 60.0           & 66.8          & 70.0          & 53.0          & \textbf{69.9} & \fcolorbox{lightgray}{lightgray}{64.7}\\
Kvasir-Sessile & 59.1          & 54.5           & 69.2          & \textbf{62.7 } & 59.9          & 45.7          & 48.9          & 48.6  & \fcolorbox{lightgray}{lightgray}{56.1}  \\
SinGAN         & 26.8          & 9.9            & 18.0          & 03.3           & \textbf{93.1} & 22.8          & 16.4          & 14.3  & \fcolorbox{lightgray}{lightgray}{25.6}  \\ 
\bottomrule
\end{tabular}}
\end{table*}
\begin{figure*}[htbp]
\centering
\subcaptionbox{on 5 polyp domain datasets
\label{fig:fig3a}}
   {\includegraphics[width=0.45\textwidth]{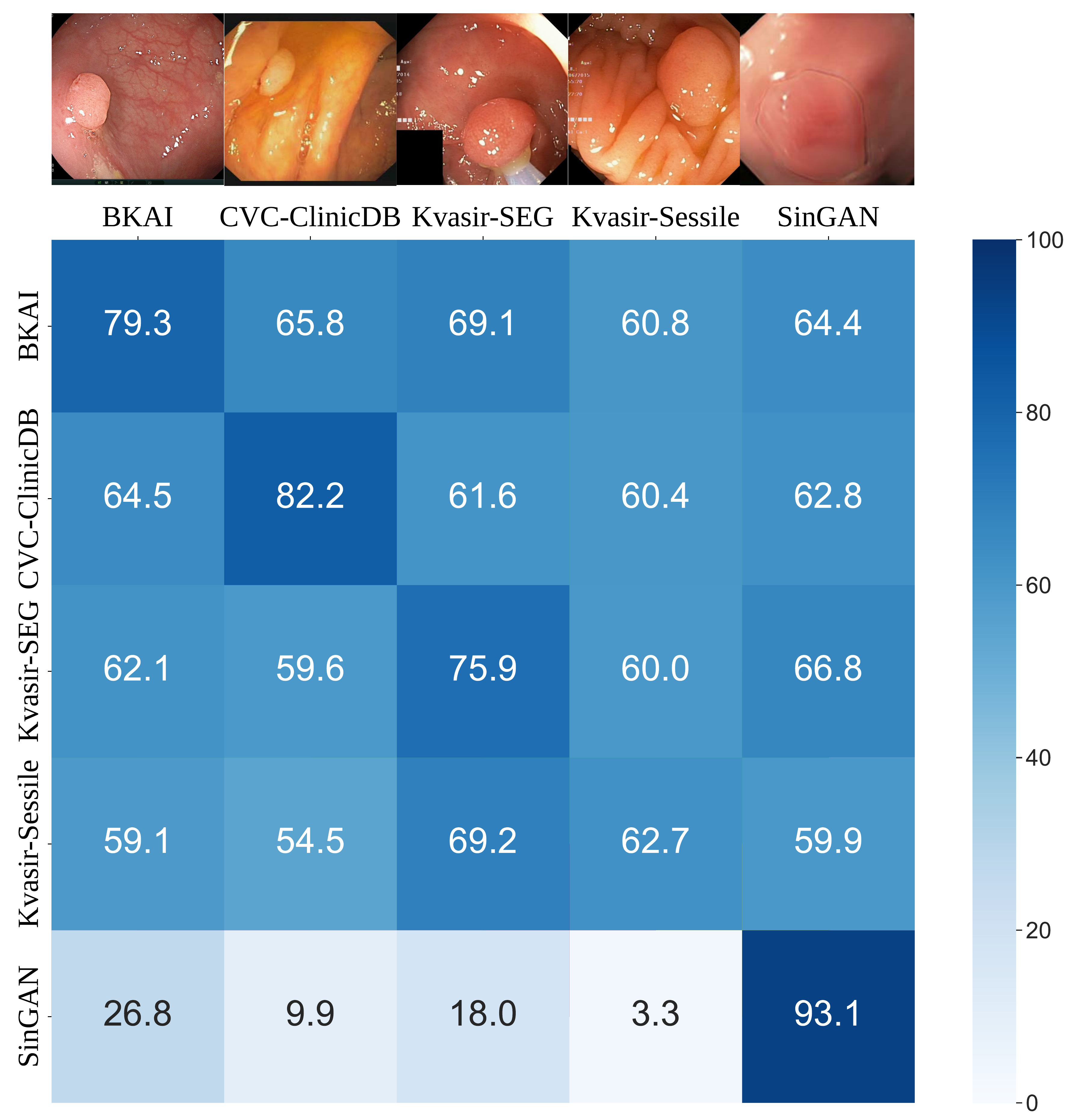}}
\subcaptionbox{on 5 medical task datasets 
\label{fig:fig3b}}
    {\includegraphics[width=0.45\textwidth]{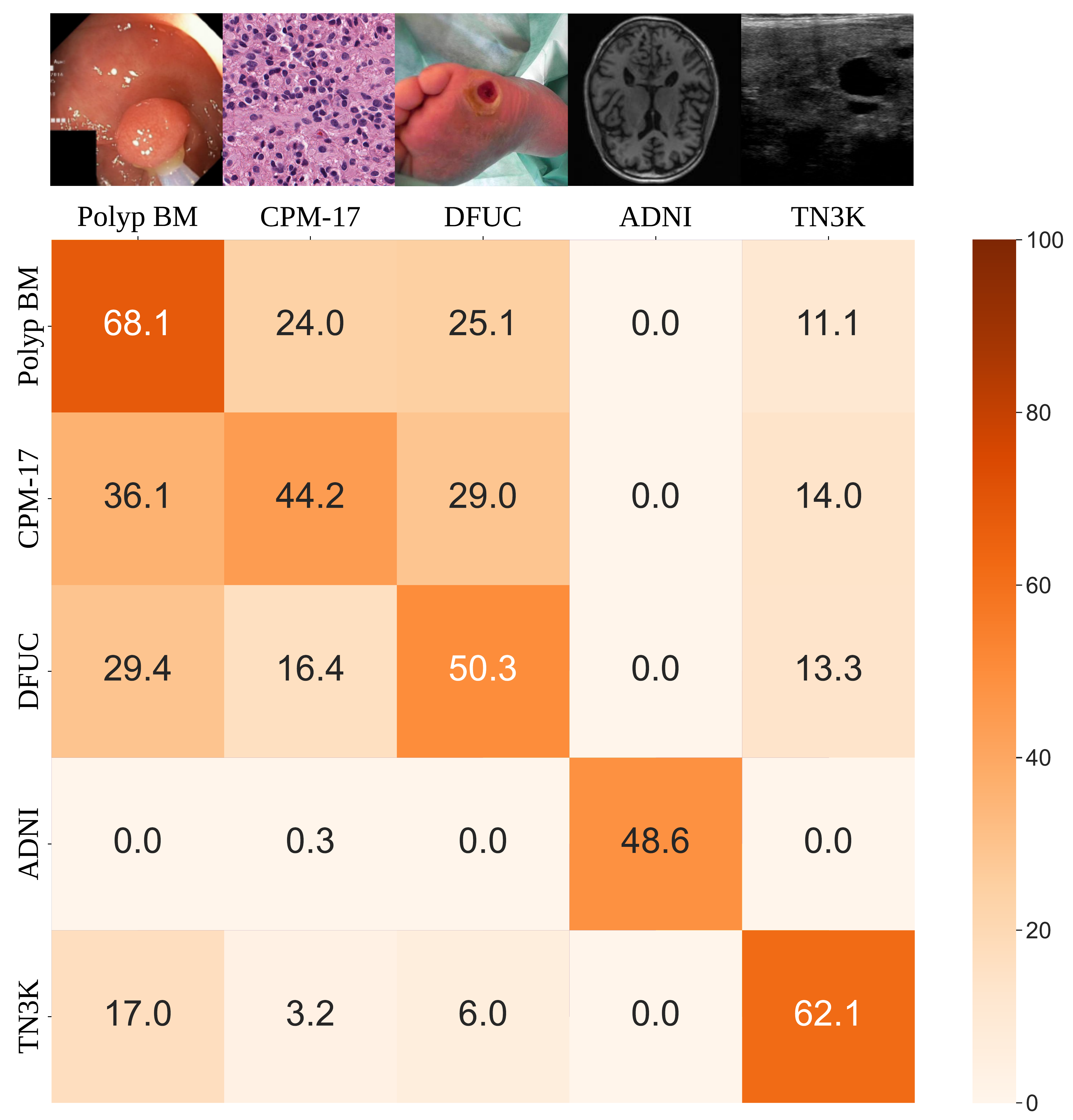}}
\caption{Cross validation results on different domains and different tasks.}
\label{fig:fig3}
\end{figure*}

\begin{table*}[ht]
\centering

\caption{Cross-task generalization performance on the medical datasets (AP\%)}
\label{tab2}
\renewcommand\arraystretch{1.5}
\setlength{\tabcolsep}{0.9mm}{
\begin{tabular}{l|ccccccccc}
\toprule
\multirow{2}{*}{\diagbox[width=9em,height=4em]{Train}{Test}} & polyp & \makecell[c]{ blood\\cell}  & \makecell[c]{ cell\\nucleus} &  \makecell[c]{diabetic\\foot ulcer} & \makecell[c]{skin\\disease} &  hippocampus & \makecell[c]{ lung\\nodule} &  \makecell[c]{pulmonary\\tuberculosis} &\makecell[c]{ thyroid\\nodule} \\ 

\cmidrule(lr){2-2} \cmidrule(lr){3-3} \cmidrule(lr){3-3} \cmidrule(lr){4-4} \cmidrule(lr){5-5} \cmidrule(lr){6-6} \cmidrule(lr){7-7} \cmidrule(lr){8-8} \cmidrule(lr){9-9} \cmidrule(lr){10-10}
             &   \makecell[c]{Polyp BM$^{\star}$}   &  BCCD          &  CPM-17 &  DFUC  &  ISIC2016 &  ADNI  &  LUNA16 &  TBX11k &  TN3K   \\ \hline
\makecell[l]{Polyp BM}       & \textbf{68.1} & 26.6          & 24.0  & 25.1 & 29.8    & 0.0 & 0.0  & 0.3  & 11.1  \\
BCCD        & 23.7          & \textbf{62.5} & 4.1  & 9.1 & 21.4    & 0.0 & 0.0  & 0.4  & 6.8  \\
CPM-17      & 36.1         & 19.6          & \textbf{44.2}  & 29.0 & 24.6    & 0.0 & 0.0  & 0.3  & 14.0  \\
DFUC        & 29.4          & 24.6          & 16.4  & \textbf{50.3} & 28.1    & 0.0 & 0.0  & 0.4  & 13.3  \\
ISIC2016    & 17.5          & 2.5           & 6.0  & 3.6 & \textbf{59.3}    & 0.0 & 0.0  & 0.3  & 5.9  \\
ADNI        & 0.0           & 0.2           & 0.3  & 0.0 & 0.1    & \textbf{48.6} & 0.0  & 0.0  & 0.0  \\
LUNA16      & 12.1          & 18.0          & 11.3  & 8.2 & 10.7    & 0.0 & \textbf{40.6}  & 0.0  & 0.1  \\
TBX11k      & 30.9          & 16.9          & 10.3  & 16.2 & 28.6    & 0.0 & 0.0  & \textbf{37.2}  & 2.3  \\
TN3K        & 17.0          & 11.7          & 3.2  & 6.0 & 20.1    & 0.0 & 0.0  & 0.9  & \textbf{62.1}  \\
\bottomrule
\multicolumn{10}{l}{\footnotesize{$^{\star}$includes CVC-300, CVC-ClinicDB, CVC-ColonDB, Kvasir, and ETIS. The results of Polyp BM are averaged over five datasets.}}
\end{tabular}}
\end{table*}
\begin{table*}[ht]
\centering
\caption{Performance in different medical domains using data incremental learning methods v.s. data fixed learning methods (AP\%)}
\label{tab3}
\renewcommand\arraystretch{1.5}
\setlength{\tabcolsep}{0.7mm}{
\begin{tabular}{lccccccccc}
\toprule
            & Kvasir-SEG & BKAI  & CVC-ClinicDB & Kvasir-Sessile & SinGAN & CVC-300 & CVC-ColonDB & ETIS  & \fcolorbox{lightgray}{lightgray}{AVG}    \\ \hline
zero-shot   & 7.1       & 4.5    & 4.1         & 4.2             & 4.2   & 6.1      & 3.2        & 1.0   & \fcolorbox{gray}{lightgray}{4.3}  \\
domain-specialized & 79.3      & 82.2   & 75.9      & 62.7 & 93.1 & 72.4 & 61.1 & 69.9 & \fcolorbox{lightgray}{lightgray}{74.6}  \\
joint (full data)  & 65.7      & 79.8   & 80.0      & 58.0            & 86.2  & 69.3     & 60.0       & 61.3  & \fcolorbox{lightgray}{lightgray}{70.0}  \\
joint (100-shot)   & 66.2      & 72.0   & 72.5        & 58.1            & 78.2  & 74.1     & 59.5       & 64.1  & \fcolorbox{lightgray}{lightgray}{68.1}  \\ \cdashline{1-10}
sequential  & 18.4      & 26.8   & 19.2        & 4.1             & \textbf{85.2}  & 26.0     & 18.7       & 19.5  & \fcolorbox{lightgray}{lightgray}{27.2}  \\
rehearsal   & \textbf{65.6} & \textbf{68.9} & \textbf{64.5} & \textbf{54.2} & 84.4 & \textbf{69.4} & \textbf{56.3} & \textbf{55.6} & \fcolorbox{lightgray}{lightgray}{\textbf{64.9}}  \\
\bottomrule
\end{tabular}}
\end{table*}
\begin{table}[ht]
\centering
\caption{Performance in different medical tasks using data incremental learning methods v.s. data fixed learning methods (AP\%)}
\label{tab4}
\renewcommand\arraystretch{1.5}
\setlength{\tabcolsep}{0.65mm}{
\begin{tabular}{lccccc}
\toprule
            & Kvasir-SEG & DFUC & TN3K  & ADNI  & \fcolorbox{lightgray}{lightgray}{AVG}    \\ \hline
zero-shot   & 8.7       & 0.8   & 1.9   & 0.0   & \fcolorbox{lightgray}{lightgray}{2.9}  \\
task-specialized & 75.9      & 50.3  & 62.1  & 48.6  & \fcolorbox{lightgray}{lightgray}{59.2}  \\
joint (full data)  & 73.6      & 49.9  & 61.0  & 47.9  & \fcolorbox{lightgray}{lightgray}{58.1}  \\
joint (100-shot)   & 64.1      & 40.2  & 45.5  & 38.6  & \fcolorbox{lightgray}{lightgray}{47.1}  \\  \cdashline{1-6}
sequential  & 1.5       & 2.2   & 16.7  & \textbf{48.8}  & \fcolorbox{lightgray}{lightgray}{17.3}  \\
rehearsal   & \textbf{57.1}     & \textbf{37.7}  & \textbf{49.2}  & 47.8  & \fcolorbox{lightgray}{lightgray}{\textbf{48.0}}  \\
\bottomrule
\end{tabular}}
\end{table}

\label{sec_experiments}
\subsection{Setup}
\paragraph{Datasets}
For domain generalization experiments, we collected a series of endoscopy image datasets for polyp detection from different domains. There are five datasets, Kvasir-SEG~\cite{Kvasir-SEG}, Kvasir-Sessile~\cite{Kvasir-Sessile}, CVC-ClinicDB~\cite{CVC-ClinicDB}, BKAI~\cite{NeoPolyp-Small}, SinGAN~\cite{SinGAN-Seg}, having train and test sets that are publicly available. Then we use them for training our models for domain generalization tasks. We also collect another three heterogeneous datasets: CVC-300~\cite{cvc-300}, CVC-ColonDB~\cite{cvc-colondb}, and ETIS~\cite{etis}, and use their test sets to evaluate our models' performance with out-of-distribution(OOD) data.

For across-tasks experiments, we focus on 8 different medical tasks: polyp detection with endocopy image dataset Kvasir-SEG~\cite{kvasir}; cell detection with cytology image dataset BCCD; cell detection with histopathology image dataset CPM-17~\cite{cpm17}; Tuberculosis detection with X-ray dataset TBX11k~\cite{tbx11k}; lung nodule detection with LUNA16~\cite{luna16}; Hippocampus area detection with ADNI~\cite{ADNI}, a dataset prepared for Alzheimer's Disease NeuroImaging research; diabetic ulcer detection with diabetic feet image dataset DFUC~\cite{dfuc2020}; thyroid nodule detection with ultrasound dataset TN3K~\cite{tn3k}

\paragraph{Implementation details}
In our specific experiments, we concentrated on the transferability and the continual learning capability of GLIP-T(C) from the natural domain to the medical domain, which is a pre-trained VLM on two natural domain datasets (Object 365 and GoldG). We carried out a large number of experiments, and in order to  ensure consistency across experiments, the design of our experiments: 1. setting the batch size to 4; 2. using Adam optimizer; 3. setting the base learning rate to $1\times10^{-4}$, setting the learning rate of the BERT text encoder to $1\times10^{-5}$; 4. setting the weight decay to 0.05; 5. freezing the bottom two layers of the image encoder. In the continual learning stage, we set the buffer size to 100.

\begin{figure*}[htbp]
\centering
\subcaptionbox{elasticity pattern
\label{fig:fig4a}}
    {\includegraphics[width=0.49\textwidth]{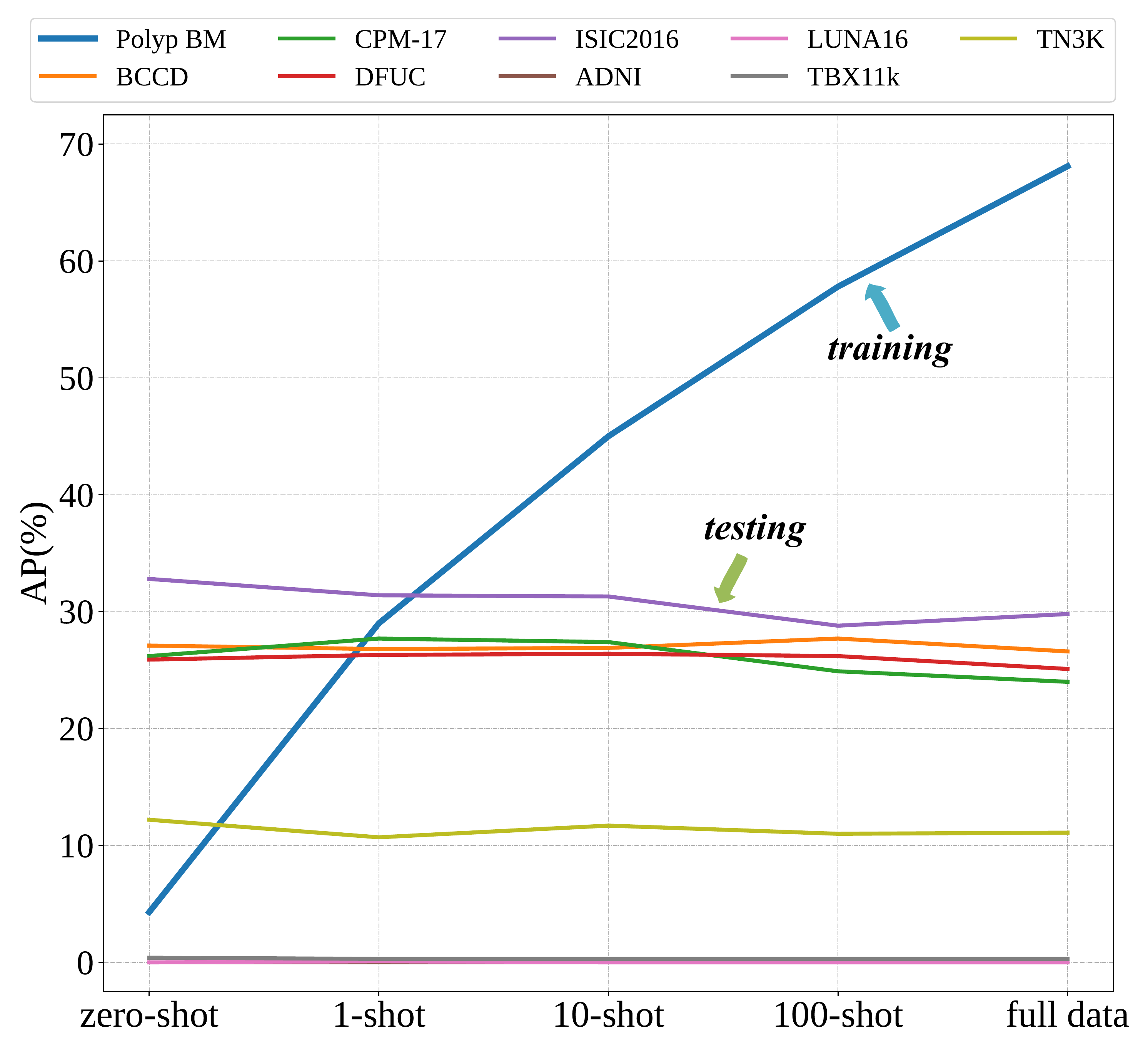}}
\subcaptionbox{forgetting pattern
\label{fig:fig4b}}
    {\includegraphics[width=0.49\textwidth]{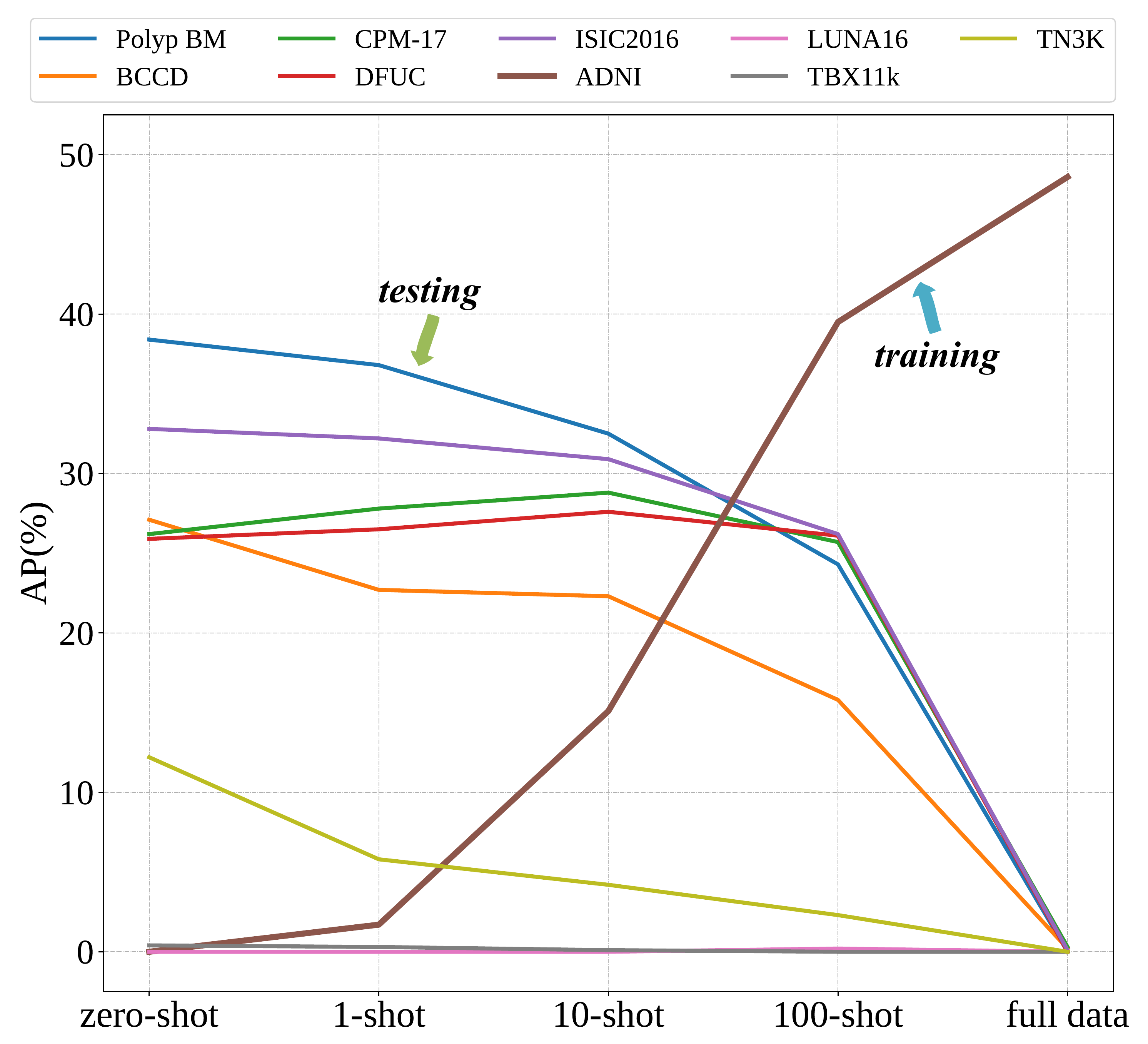}}
\caption{Two typical generalization performance patterns: elasticity v.s. forgetting}
\label{fig:fig4}
\end{figure*}

\subsection{Generalization Performance Analysis}
\subsubsection{Domain/task specialized models have poor generalization performance} As mentioned before, domain/task-specialized models can not satisfy the need of generalization capabilities in clinical use cases. In this section, we will use experimental results to show such incompetence of the domain/task-specialized models. Table~\ref{tab1} and Table~\ref{tab2} record the domain/task-specialized models' performance on domain-generalization and task-generalization tasks, respectively. Fig.~\ref{fig:fig3a} and Fig.~\ref{fig:fig3b} are the visualized heatmap figures of these two tables. As we explained before, each row in either Table~\ref{tab1} or Table~\ref{tab2} represents a model trained with the according data and its performance on the test sets with all other domains/tasks. As illustrated in those two tables, we can conclude that the domain/task-specialized models can achieve strong performance on the test data from its own domain/task, but their performance on other datasets is not satisfying enough. This pattern can be clearly observed from the visualized heatmap Fig.~\ref{fig:fig3a} and Fig.~\ref{fig:fig3b}. The diagonal numbers in Fig.~\ref{fig:fig3a} and Fig.~\ref{fig:fig3b} are clearly larger than those in other positions, and these tell us the domain/task gaps are significant. Compared with Fig.~\ref{fig:fig3a}, the task gaps are more evident in Fig.~\ref{fig:fig3b} since the performance differences are much larger. And this observation suggests that domain/task-specialized models are definitely not the correct path toward the foundation models because we need models that can handle different downstream tasks simultaneously.

\subsubsection{General knowledge acqusition: two patterns}
We also visualized two typical generalization performance patterns in Fig.~\ref{fig:fig4}. As illustrated in Fig.~\ref{fig:fig4a}, the generalization performance pattern on various datasets of the model trained with the Kvasir-SEG dataset is called elasticity in continual learning. As one can tell, the model's test results on other datasets keep the same, while the performance on its own test set is increasing. This pattern can be interpreted from two perspectives. Firstly, the model in the left figure doesn't suffer from the catastrophic problem, since it doesn't lose two much generalization capability during training. Secondly, the model is not learning any transferable knowledge from this dataset. On the other hand, the model, which is trained with the ADNI dataset, has another performance pattern on the test sets from different tasks. The pattern in Fig.~\ref{fig:fig4b} illustrates that the task-generalization capability is actually decreasing during training with more data. The reason for such a phenomenon, as we guess, is probably because of the relatively large difficulty of the ADNI dataset's detection task, compared to other datasets. This observation shows that the model can lose generalization capability during training, which suggests that we need an approach to prevent forgetting common knowledge.

Therefore, in the later sections, we will show that joint learning and continual learning show more promising results for cross-domain and cross-task experiments.

\begin{figure*}[htbp]
\centering
\subcaptionbox{on cross-domain datasets
\label{fig:fig5a}}
    {\includegraphics[width=0.49\textwidth]{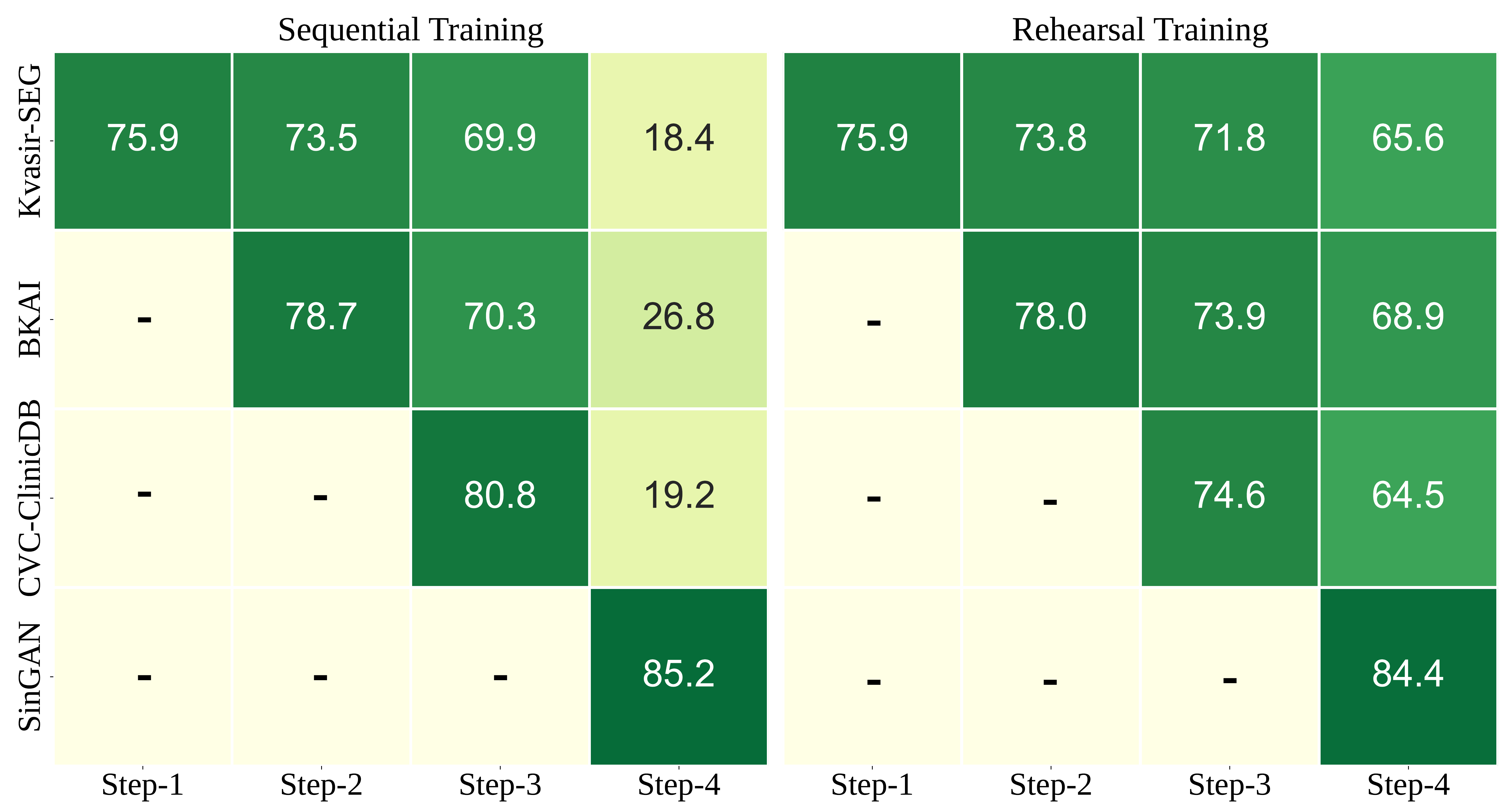}}
\subcaptionbox{on cross-task datasets
\label{fig:fig5b}}
    {\includegraphics[width=0.49\textwidth]{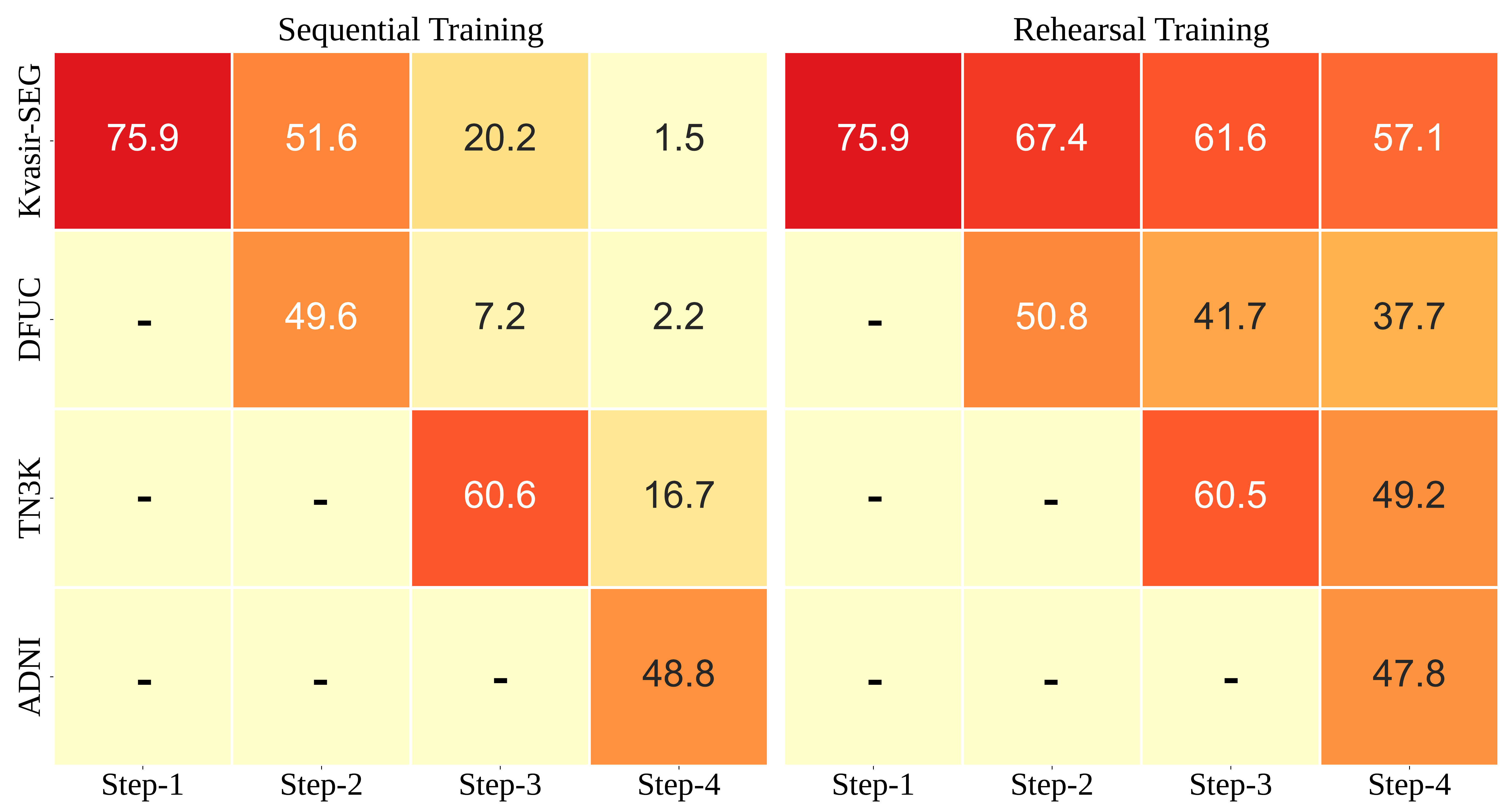}}
\caption{The performance of rehearsal training v.s. sequential training on cross-domain/cross-task datasets}
\label{fig:fig5}
\end{figure*}

\subsubsection{Joint learning sharply increases the generalization performance}
For joint learning, we use data from different domains/tasks to train the model simultaneously, and we test the pre-trained model on the test sets from various domains/tasks. Table~\ref{tab3} exhibits the generalization performance of the model that was jointly trained on all the available polyp train data we mentioned before. Row 3 and 4 in Table~\ref{tab3}, respectively show the results of the model jointly trained under full data and 100 shots setting. And the second row records the test result of the model that was trained and tested on the same data domain, which is the result on the diagonal line in Fig.~\ref{fig:fig3}. And the zero-shot line shows the benchmark results of directly using the not finetuned VLM to do the tests by giving them the class name as text prompts.
From the first four rows of Table~\ref{tab3}, we can conclude that joint training does boost the model's generalization capability, but the results are still relatively lower than those of the upper bound one average. This pattern is understandable since the models have to sacrifice some domain-specialized performance to exchange the increase of generalization capability.
We also notice that the though the joint learning model with 100 shots has a relatively weak performance on average, the few-shot model achieves better results on the OOD test data ( CVC-300, CVC-ColonDB, and ETIS). This interesting phenomenon gives us some inspiration. Is it possible to train a joint learning model without using the full-size dataset and losing no generalization capability? As we are sampling the 100 shots data randomly, we think that, with a more advanced sampling strategy, a few-shot joint learning model may achieve satisfying results. And we there are some data which are more representative compared to their counterparts. However, we will leave this discussion in our future works due to space limitations. We believe this direction is worth exploring since a few-shot joint learning model is much more realistic and practical in the real world.

\subsubsection{Basic sequential learning methods suffering from catastrophic forgetting}
We now present the experiment results of the models trained with the continual learning paradigm. However, before we step into the detail, we want to remind the readers that continual learning is a more practical option in many clinical use cases. Compared with the joint learning paradigm above, continual learning won't require a large dataset to train simultaneously and doesn't need to train the model every time we add unseen data from a new domain/task. In Table~\ref{tab3} and Table~\ref{tab4}, we can see the last two rows represent the results of the model trained with sequential and rehearsal methods respectively. 
The first conclusion we can make is that the sequential learning method badly suffers from the catastrophic forgetting problem and receives unsuccessful results regarding generalization capability. Fig.~\ref{fig:fig5} illustrates the detailed training process of the sequential learning model and rehearsal learning model. Let's take Fig.~\ref{fig:fig5b} as an example. This figure displays the intermediate results from the training process with the cross-domain scenario. The left heatmap shows the training process of sequential learning, and the right heatmap represents the training process of rehearsal learning. Each column of the heatmap represents the cross-task results of the model at $step_i$, and a step means the whole training process on a dataset for a specific task. The order of adding the dataset is the same as the dataset name on the y-axis from top to bottom. Thus, the results in the diagonal lines are generated by the continual learning model that is currently learning and testing with the same dataset. After understanding the above facts, we can clearly see that the sequential learning model forgets most of the knowledge acquired after two steps. As illustrated in the left heatmap of Fig.~\ref{fig:fig5b}, the result of Kvasir SEG dataset sharply plunged from 75.9\% at step 1 to 1.5\% at step 4.

\subsubsection{Rehearsal learning with replay buffer can prevent forgetting}
Compared to the unsatisfactory generalization performance of the sequential learning models, the rehearsal learning models achieve much better results in cross-domain/task tests, while the only difference is the introduction of a replay buffer. In fact, the rehearsal learning method obtains a better average result than the joint-100 method, meaning rehearsal learning has a promising generalization capability. Again, we want to remind the readers that we only use the most straightforward sampling strategy, random sampling, to implement rehearsal learning. We believe we could achieve better performance by adopting another sampling strategy, but we will leave this exploration to our feature work. From the results in Table~\ref{tab3} and Table~\ref{tab4}, we can conclude that the introduction of the replay buffer significantly increases the cross-domain performance. For example, the average cross-task result of the sequential learning model is 17.3\%, and by introducing the replay buffer, this number comes to 48.0\% of the rehearsal model. In Fig.~\ref{fig:fig5b}, from the right heatmap, we can see the result of the Kvasir SEG dataset only moderately dropped from 75.9\% at step 1 to 57.1\% at step 4. This drop is much more acceptable compared to the former one. And this figure again supports our conclusion before. Catastrophic forgetting of the models happens sooner than we thought, in only two or three steps. Furthermore, the rehearsal learning paradigm is both cost-efficient and data-efficient to alleviate the forgetting problem. In the end, we can conclude that the catastrophic forgetting problem is unavoidable for adapting pre-trained models to a target domain. Thus, we must invest more elaboration in preventing such forgetting in practice.

\section{Conclusion}
\label{sec_conclusion}
In this work, we first highlight the rise of large-size foundation models and their superiority in many fields. Then we envision a future where foundation models can also serve as a general-purpose system in medical image tasks. Therefore, we discuss and evaluate several training paradigms in terms of generalization capability for cross-domain and cross-task tests. We observed some interesting patterns from all the experiments we conducted and drew the following conclusion: First, the domain/task-specialized models have little generalization capability; Second, the joint learning method can increase the models' generalization performance but require large-size heterogeneous data; Third, compared to sequential learning, rehearsal learning can better alleviate the forgetting problem of the pre-trained models. Except for the above conclusions, we also leave some open questions for future works moving toward the medical foundation models. The first question is whether the joint learning method requires full data to train the model. Is there a strategy to sample more representative data from each dataset? The second question is about the rehearsal learning paradigm. Could a more advanced replay method increase the rehearsal learning models' generalization capability and prevent the forgetting problem? Ultimately, we hope this work can provide some practical guidance for future researchers.

{\small
\bibliographystyle{ieee_fullname}

\bibliography{Ref.bib}

\begin{thebibliography}{10}\itemsep=-1pt

\bibitem{bachmann2022multimae}
Roman Bachmann, David Mizrahi, Andrei Atanov, and Amir Zamir.
\newblock Multimae: Multi-modal multi-task masked autoencoders.
\newblock In {\em Computer Vision--ECCV 2022: 17th European Conference, Tel
  Aviv, Israel, October 23--27, 2022, Proceedings, Part XXXVII}, pages
  348--367. Springer, 2022.

\bibitem{CVC-ClinicDB}
Jorge Bernal, F~Javier S{\'a}nchez, Gloria Fern{\'a}ndez-Esparrach, Debora Gil,
  Cristina Rodr{\'\i}guez, and Fernando Vilari{\~n}o.
\newblock Wm-dova maps for accurate polyp highlighting in colonoscopy:
  Validation vs. saliency maps from physicians.
\newblock {\em Computerized medical imaging and graphics}, 43:99--111, 2015.

\bibitem{ADNI}
Marina Boccardi, Martina Bocchetta, F{\'e}lix~C Morency, D~Louis Collins,
  Masami Nishikawa, Rossana Ganzola, Michel~J Grothe, Dominik Wolf, Alberto
  Redolfi, Michela Pievani, et~al.
\newblock Training labels for hippocampal segmentation based on the eadc-adni
  harmonized hippocampal protocol.
\newblock {\em Alzheimer's \& Dementia}, 11(2):175--183, 2015.

\bibitem{bommasani2021opportunities}
Rishi Bommasani, Drew~A Hudson, Ehsan Adeli, Russ Altman, Simran Arora, Sydney
  von Arx, Michael~S Bernstein, Jeannette Bohg, Antoine Bosselut, Emma
  Brunskill, et~al.
\newblock On the opportunities and risks of foundation models.
\newblock {\em arXiv preprint arXiv:2108.07258}, 2021.

\bibitem{dfuc2020}
Bill Cassidy, Neil~D Reeves, Joseph~M Pappachan, David Gillespie, Claire
  O’Shea, Satyan Rajbhandari, Arun~G Maiya, Eibe Frank, Andrew~JM Boulton,
  David~G Armstrong, et~al.
\newblock The dfuc 2020 dataset: Analysis towards diabetic foot ulcer
  detection.
\newblock {\em touchREVIEWS in Endocrinology}, 17(1):5, 2021.

\bibitem{devlin2018bert}
Jacob Devlin, Ming-Wei Chang, Kenton Lee, and Kristina Toutanova.
\newblock Bert: Pre-training of deep bidirectional transformers for language
  understanding.
\newblock {\em arXiv preprint arXiv:1810.04805}, 2018.

\bibitem{dong2022bootstrapped}
Xiaoyi Dong, Jianmin Bao, Ting Zhang, Dongdong Chen, Weiming Zhang, Lu Yuan,
  Dong Chen, Fang Wen, and Nenghai Yu.
\newblock Bootstrapped masked autoencoders for vision bert pretraining.
\newblock In {\em Computer Vision--ECCV 2022: 17th European Conference, Tel
  Aviv, Israel, October 23--27, 2022, Proceedings, Part XXX}, pages 247--264.
  Springer, 2022.

\bibitem{douillard2022dytox}
Arthur Douillard, Alexandre Ram{\'e}, Guillaume Couairon, and Matthieu Cord.
\newblock Dytox: Transformers for continual learning with dynamic token
  expansion.
\newblock In {\em Proceedings of the IEEE/CVF Conference on Computer Vision and
  Pattern Recognition}, pages 9285--9295, 2022.

\bibitem{french1999catastrophic}
Robert~M French.
\newblock Catastrophic forgetting in connectionist networks.
\newblock {\em Trends in cognitive sciences}, 3(4):128--135, 1999.

\bibitem{tn3k}
Haifan Gong, Guanqi Chen, Ranran Wang, Xiang Xie, Mingzhi Mao, Yizhou Yu, Fei
  Chen, and Guanbin Li.
\newblock Multi-task learning for thyroid nodule segmentation with thyroid
  region prior.
\newblock In {\em 2021 IEEE 18th International Symposium on Biomedical Imaging
  (ISBI)}, pages 257--261. IEEE, 2021.

\bibitem{he2022masked}
Kaiming He, Xinlei Chen, Saining Xie, Yanghao Li, Piotr Doll{\'a}r, and Ross
  Girshick.
\newblock Masked autoencoders are scalable vision learners.
\newblock In {\em Proceedings of the IEEE/CVF Conference on Computer Vision and
  Pattern Recognition}, pages 16000--16009, 2022.

\bibitem{Kvasir-Sessile}
Debesh Jha, Pia~H Smedsrud, Dag Johansen, Thomas de Lange, H{\aa}vard~D
  Johansen, P{\aa}l Halvorsen, and Michael~A Riegler.
\newblock A comprehensive study on colorectal polyp segmentation with
  resunet++, conditional random field and test-time augmentation.
\newblock {\em IEEE journal of biomedical and health informatics},
  25(6):2029--2040, 2021.

\bibitem{Kvasir-SEG}
Debesh Jha, Pia~H Smedsrud, Michael~A Riegler, P{\aa}l Halvorsen, Thomas de
  Lange, Dag Johansen, and H{\aa}vard~D Johansen.
\newblock Kvasir-seg: A segmented polyp dataset.
\newblock In {\em MultiMedia Modeling: 26th International Conference, MMM 2020,
  Daejeon, South Korea, January 5--8, 2020, Proceedings, Part II 26}, pages
  451--462. Springer, 2020.

\bibitem{kvasir}
Debesh Jha, Pia~H Smedsrud, Michael~A Riegler, P{\aa}l Halvorsen, Thomas~de
  Lange, Dag Johansen, and H{\aa}vard~D Johansen.
\newblock Kvasir-seg: A segmented polyp dataset.
\newblock In {\em International Conference on Multimedia Modeling}, pages
  451--462. Springer, 2020.

\bibitem{ALIGN}
Chao Jia, Yinfei Yang, Ye Xia, Yi-Ting Chen, Zarana Parekh, Hieu Pham, Quoc Le,
  Yun-Hsuan Sung, Zhen Li, and Tom Duerig.
\newblock Scaling up visual and vision-language representation learning with
  noisy text supervision.
\newblock In {\em International Conference on Machine Learning}, pages
  4904--4916. PMLR, 2021.

\bibitem{Lao2021ATC}
Qicheng Lao, Xiangxi Jiang, Mohammad Havaei, and Yoshua Bengio.
\newblock A two-stream continual learning system with variational
  domain-agnostic feature replay.
\newblock {\em IEEE Transactions on Neural Networks and Learning Systems},
  33:4466--4478, 2021.

\bibitem{Lao2020FoCLFC}
Qicheng Lao, Mehrzad Mortazavi, Marzieh~S. Tahaei, Francis Dutil, T. Fevens,
  and Mohammad Havaei.
\newblock Focl: Feature-oriented continual learning for generative models.
\newblock {\em Pattern Recognit.}, 120:108127, 2020.

\bibitem{lee2019patentbert}
Jieh-Sheng Lee and Jieh Hsiang.
\newblock Patentbert: Patent classification with fine-tuning a pre-trained bert
  model.
\newblock {\em arXiv preprint arXiv:1906.02124}, 2019.

\bibitem{li2022grounded}
Liunian~Harold Li, Pengchuan Zhang, Haotian Zhang, Jianwei Yang, Chunyuan Li,
  Yiwu Zhong, Lijuan Wang, Lu Yuan, Lei Zhang, Jenq-Neng Hwang, et~al.
\newblock Grounded language-image pre-training.
\newblock In {\em Proceedings of the IEEE/CVF Conference on Computer Vision and
  Pattern Recognition}, pages 10965--10975, 2022.

\bibitem{tbx11k}
Yun Liu, Yu-Huan Wu, Yunfeng Ban, Huifang Wang, and Ming-Ming Cheng.
\newblock Rethinking computer-aided tuberculosis diagnosis.
\newblock In {\em Proceedings of the IEEE/CVF conference on computer vision and
  pattern recognition}, pages 2646--2655, 2020.

\bibitem{NeoPolyp-Small}
Phan Ngoc~Lan, Nguyen~Sy An, Dao~Viet Hang, Dao~Van Long, Tran~Quang Trung,
  Nguyen~Thi Thuy, and Dinh~Viet Sang.
\newblock Neounet: Towards accurate colon polyp segmentation and neoplasm
  detection.
\newblock In {\em Advances in Visual Computing: 16th International Symposium,
  ISVC 2021, Virtual Event, October 4-6, 2021, Proceedings, Part II}, pages
  15--28. Springer, 2021.

\bibitem{lama}
Fabio Petroni, Tim Rockt{\"a}schel, Patrick Lewis, Anton Bakhtin, Yuxiang Wu,
  Alexander~H Miller, and Sebastian Riedel.
\newblock Language models as knowledge bases?
\newblock {\em arXiv preprint arXiv:1909.01066}, 2019.

\bibitem{qin2022medical}
Ziyuan Qin, Huahui Yi, Qicheng Lao, and Kang Li.
\newblock Medical image understanding with pretrained vision language models: A
  comprehensive study.
\newblock {\em arXiv preprint arXiv:2209.15517}, 2022.

\bibitem{CLIP}
Alec Radford, Jong~Wook Kim, Chris Hallacy, Aditya Ramesh, Gabriel Goh,
  Sandhini Agarwal, Girish Sastry, Amanda Askell, Pamela Mishkin, Jack Clark,
  et~al.
\newblock Learning transferable visual models from natural language
  supervision.
\newblock In {\em International Conference on Machine Learning}, pages
  8748--8763. PMLR, 2021.

\bibitem{Schick2020ExploitingCF}
Timo Schick and Hinrich Sch{\"u}tze.
\newblock Exploiting cloze-questions for few-shot text classification and
  natural language inference.
\newblock In {\em Conference of the European Chapter of the Association for
  Computational Linguistics}, 2020.

\bibitem{luna16}
Arnaud Arindra~Adiyoso Setio, Alberto Traverso, Thomas De~Bel, Moira~SN Berens,
  Cas Van Den~Bogaard, Piergiorgio Cerello, Hao Chen, Qi Dou, Maria~Evelina
  Fantacci, Bram Geurts, et~al.
\newblock Validation, comparison, and combination of algorithms for automatic
  detection of pulmonary nodules in computed tomography images: the luna16
  challenge.
\newblock {\em Medical image analysis}, 42:1--13, 2017.

\bibitem{etis}
Juan Silva, Aymeric Histace, Olivier Romain, Xavier Dray, and Bertrand Granado.
\newblock Toward embedded detection of polyps in wce images for early diagnosis
  of colorectal cancer.
\newblock {\em International journal of computer assisted radiology and
  surgery}, 9(2):283--293, 2014.

\bibitem{HaoyuSong2022CLIPMA}
Haoyu Song, Li Dong, Wei-Nan Zhang, Ting Liu, and Furu Wei.
\newblock Clip models are few-shot learners: Empirical studies on vqa and
  visual entailment.
\newblock {\em arXiv preprint arXiv:2203.07190}, 2022.

\bibitem{cvc-colondb}
Nima Tajbakhsh, Suryakanth~R Gurudu, and Jianming Liang.
\newblock Automated polyp detection in colonoscopy videos using shape and
  context information.
\newblock {\em IEEE transactions on medical imaging}, 35(2):630--644, 2015.

\bibitem{SinGAN-Seg}
Vajira Thambawita, Pegah Salehi, Sajad~Amouei Sheshkal, Steven~A Hicks, Hugo~L
  Hammer, Sravanthi Parasa, Thomas~de Lange, P{\aa}l Halvorsen, and Michael~A
  Riegler.
\newblock Singan-seg: Synthetic training data generation for medical image
  segmentation.
\newblock {\em PloS one}, 17(5):e0267976, 2022.

\bibitem{thengane2022clip}
Vishal Thengane, Salman Khan, Munawar Hayat, and Fahad Khan.
\newblock Clip model is an efficient continual learner.
\newblock {\em arXiv preprint arXiv:2210.03114}, 2022.

\bibitem{tompson2014joint}
Jonathan~J Tompson, Arjun Jain, Yann LeCun, and Christoph Bregler.
\newblock Joint training of a convolutional network and a graphical model for
  human pose estimation.
\newblock {\em Advances in neural information processing systems}, 27, 2014.

\bibitem{van2022three}
Gido~M van~de Ven, Tinne Tuytelaars, and Andreas~S Tolias.
\newblock Three types of incremental learning.
\newblock {\em Nature Machine Intelligence}, pages 1--13, 2022.

\bibitem{cvc-300}
David V{\'a}zquez, Jorge Bernal, F~Javier S{\'a}nchez, Gloria
  Fern{\'a}ndez-Esparrach, Antonio~M L{\'o}pez, Adriana Romero, Michal
  Drozdzal, and Aaron Courville.
\newblock A benchmark for endoluminal scene segmentation of colonoscopy images.
\newblock {\em Journal of healthcare engineering}, 2017, 2017.

\bibitem{cpm17}
Quoc~Dang Vu, Simon Graham, Tahsin Kurc, Minh Nguyen~Nhat To, Muhammad Shaban,
  Talha Qaiser, Navid~Alemi Koohbanani, Syed~Ali Khurram, Jayashree
  Kalpathy-Cramer, Tianhao Zhao, et~al.
\newblock Methods for segmentation and classification of digital microscopy
  tissue images.
\newblock {\em Frontiers in bioengineering and biotechnology}, page~53, 2019.

\bibitem{OFA}
Peng Wang, An Yang, Rui Men, Junyang Lin, Shuai Bai, Zhikang Li, Jianxin Ma,
  Chang Zhou, Jingren Zhou, and Hongxia Yang.
\newblock Unifying architectures, tasks, and modalities through a simple
  sequence-to-sequence learning framework.
\newblock {\em arXiv preprint arXiv:2202.03052}, 2022.

\bibitem{weng2020acquiring}
Rongxiang Weng, Heng Yu, Shujian Huang, Shanbo Cheng, and Weihua Luo.
\newblock Acquiring knowledge from pre-trained model to neural machine
  translation.
\newblock In {\em Proceedings of the AAAI conference on artificial
  intelligence}, volume~34, pages 9266--9273, 2020.

\bibitem{wojcik2022foundation}
Malwina~Anna W{\'o}jcik.
\newblock Foundation models in healthcare: Opportunities, biases and regulatory
  prospects in europe.
\newblock In {\em Electronic Government and the Information Systems
  Perspective: 11th International Conference, EGOVIS 2022, Vienna, Austria,
  August 22--24, 2022, Proceedings}, pages 32--46. Springer, 2022.

\bibitem{JianweiYang2022UnifiedCL}
Jianwei Yang, Chunyuan Li, Pengchuan Zhang, Bin Xiao, Ce Liu, Lu Yuan, and
  Jianfeng Gao.
\newblock Unified contrastive learning in image-text-label space.
\newblock In {\em Proceedings of the IEEE/CVF Conference on Computer Vision and
  Pattern Recognition}, pages 19163--19173, 2022.

\bibitem{Zhang2019JointLO}
Zaiwei Zhang, Xiangru Huang, Qixing Huang, Xiao Zhang, and Yuan Li.
\newblock Joint learning of neural networks via iterative reweighted least
  squares.
\newblock In {\em Cvpr workshops}, pages 18--26, 2019.

\end{thebibliography}
}

\end{document}